%% file: main.tex
\documentclass[10pt,twocolumn,letterpaper]{article}

\usepackage[pagenumbers]{cvpr}  

\usepackage{cuted}
\usepackage{graphicx}
\usepackage{capt-of}
\usepackage{xcolor}
\usepackage{xspace}
\usepackage{amsmath,amssymb,amsbsy,amsfonts,dsfont,pifont,bm,bbm,mathrsfs,mathtools,nicefrac}
\usepackage{float,wrapfig}
\usepackage{booktabs,multirow,adjustbox,diagbox,threeparttable,tabularray}
\usepackage{algorithm,algpseudocode,listings}
\usepackage[accsupp]{axessibility}
\usepackage{verbatim}
\usepackage{enumitem}
\definecolor{citeblue}{RGB}{48,111,186}
\definecolor{citecolor}{HTML}{0071BC}
\definecolor{linkcolor}{HTML}{ED1C24}
\usepackage[pagebackref=false,breaklinks=true,colorlinks=true,citecolor=citeblue,bookmarks=false]{hyperref}
\usepackage{graphicx}

\usepackage{cleveref}  
\crefname{section}{Sec.}{Secs.}
\Crefname{section}{Section}{Sections}
\crefname{table}{Tab.}{Tabs.}
\Crefname{table}{Table}{Tables}
\crefname{figure}{Fig.}{Figs.}
\Crefname{figure}{Figure}{Figures}
\crefname{equation}{Eq.}{Eqs.}
\Crefname{equation}{Equation}{Equations}
\newcommand{\tocite}[1]{{\color{red}[TO CITE]}}
\newcommand{\method}{\textcolor{black}{\mbox{\texttt{CoDeF}}}\xspace}

\def\cvprPaperID{****}
\def\confName{CVPR}
\def\confYear{2023}

\newcommand\nonumfootnote[1]{%
\begingroup%
    \renewcommand\thefootnote{}\footnote{\hspace{-3.7pt}#1}%
    \addtocounter{footnote}{-1}%
\endgroup%
}

\begin{document}

\title{CoDeF: Content Deformation Fields for Temporally Consistent Video Processing}

\author{
    Hao Ouyang\textsuperscript{1,2*} \quad
    Qiuyu Wang\textsuperscript{2*} \quad
    Yuxi Xiao\textsuperscript{2,3*} \quad
    Qingyan Bai\textsuperscript{1,2} \quad
    Juntao Zhang\textsuperscript{1} \\[2pt]
    Kecheng Zheng\textsuperscript{2} \quad
    Xiaowei Zhou\textsuperscript{3} \quad
    Qifeng Chen\textsuperscript{1$\dagger$} \quad
    Yujun Shen\textsuperscript{2$\dagger$} \\[5pt]
    $^1$HKUST \qquad
    $^2$Ant Group \qquad
    $^3$CAD\&CG, ZJU \\[8pt]
}

\twocolumn[{
\renewcommand\twocolumn[1][]{#1}
\maketitle
\begin{center}
    \vspace{-15pt}
    \includegraphics[width=1.0\linewidth]{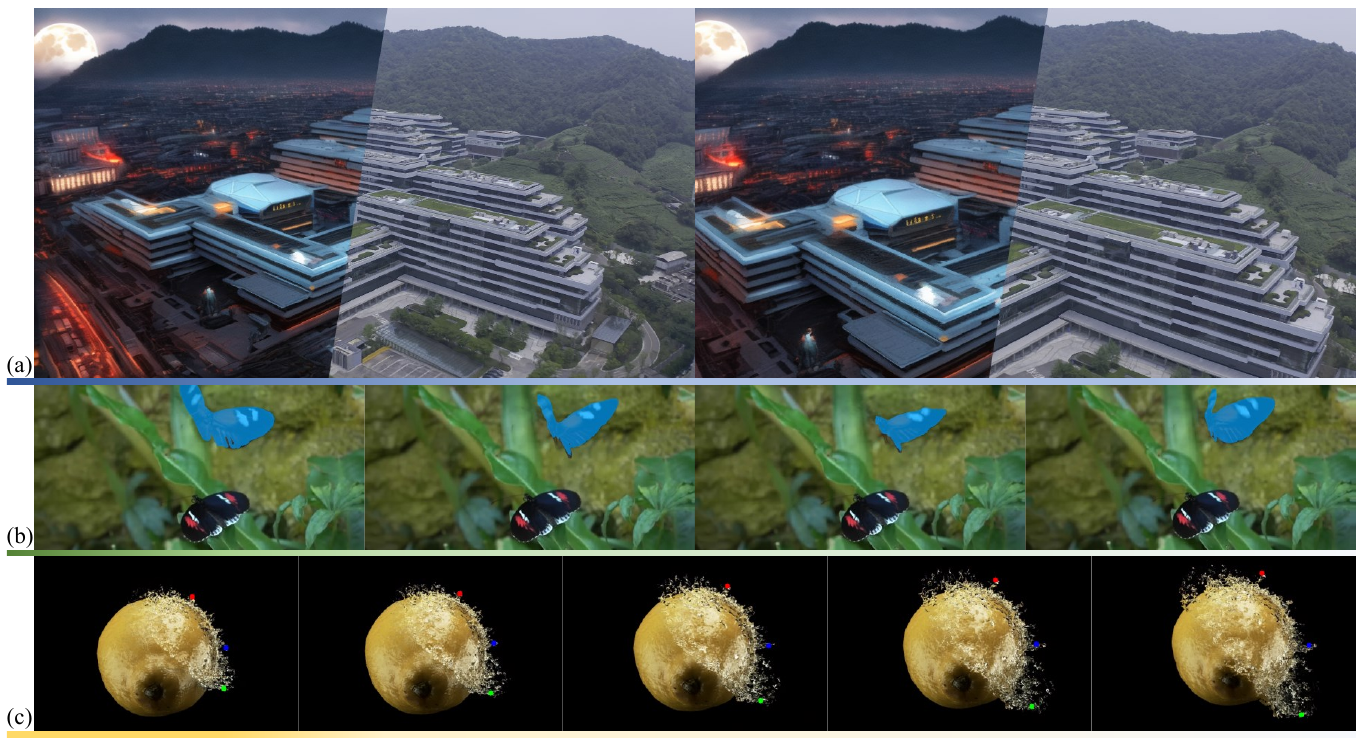}
    \vspace{-17pt}
    \captionsetup{type=figure}
    \caption{%
        \textbf{Versatile applications} of \method including (a) text-guided video-to-video translation (left half: translated frames, right half: input frames),  (b) video object tracking, and (c) video keypoint tracking.
        It is noteworthy that, with the proposed type of video representation, we manage to directly \textit{lift} image algorithms for video processing \textit{without any tuning} on videos.
    }
    \label{fig:teaser}
    \vspace{7pt}
\end{center}
}]

\input{sections/0.abs.tex}
\input{sections/1.intro.tex}

\input{sections/2.related.tex}

\input{sections/3.method.tex}

\input{sections/4.exp.tex}
\input{sections/5.conclusion.tex}
\input{sections/6.ref.tex}

\end{document}

%% file: sections/0.abs.tex
\begin{abstract}

We present the content deformation field (\method) as a new type of video representation, which consists of a \textbf{canonical content field} aggregating the static contents in the entire video and a \textbf{temporal deformation field} recording the transformations from the canonical image (\textit{i.e.}, rendered from the canonical content field) to each individual frame along the time axis.
Given a target video, these two fields are jointly optimized to reconstruct it through a carefully tailored rendering pipeline.
We advisedly introduce some regularizations into the optimization process, urging the canonical content field to inherit semantics (\textit{e.g.}, the object shape) from the video.
With such a design, \method naturally supports lifting image algorithms for video processing, in the sense that one can apply an image algorithm to the canonical image and effortlessly propagate the outcomes to the entire video with the aid of the temporal deformation field.
We experimentally show that \method is able to lift image-to-image translation to video-to-video translation and lift keypoint detection to keypoint tracking without any training.
More importantly, thanks to our lifting strategy that deploys the algorithms on only one image, we achieve superior cross-frame consistency in processed videos compared to existing video-to-video translation approaches, and even manage to track non-rigid objects like water and smog.
Project page can be found \href{https://qiuyu96.github.io/CoDeF/}{here}.
\nonumfootnote{* Equal contribution \qquad $\dagger$ Corresponding author}

\end{abstract}
\vspace{-30pt}

%% file: sections/1.intro.tex
\section{Introduction}\label{sec:intro}
The field of image processing has witnessed remarkable advancements, largely attributable to the power of generative models trained on extensive datasets, yielding exceptional quality and precision. However, the processing of video content has not achieved comparable progress. One challenge lies in maintaining high temporal consistency, a task complicated by the inherent randomness of neural networks. Another challenge arises from the nature of video datasets themselves, which often include textures of inferior quality compared to their image counterparts and necessitate greater computational resources. Consequently, the  quality of video-based algorithms significantly lags behind those focused on images. This contrast prompts a question: \textit{is it feasible to represent video in the form of an image to seamlessly apply established image algorithms to video content with high temporal consistency}?

In pursuit of this objective, researchers have suggested the generation of video mosaics from dynamic videos~\cite{rav2008unwrap, shade1998layered} in the era preceding deep learning, and the utilization of a neural layered image atlas~\cite{kasten2021layered, lu2020layered, ye2022deformable} subsequent to the proposal of implicit neural representations. Nonetheless, these methods exhibit two principal deficiencies. First, the capacity of these representations, particularly in faithfully reconstructing intricate details within a video, is restricted. Often, the reconstructed video overlooks subtle motion details, such as blinking eyes or slight smiles. The second limitation pertains to the typically distorted nature of the estimated atlas, which consequently suffers from impaired semantic information. Existing image processing algorithms, therefore, do not perform optimally as the estimated atlas lacks sufficient naturalness.

We propose a novel approach to video representation that utilizes a 2D hash-based image field coupled with a 3D hash-based temporal deformation field. The incorporation of multi-resolution hash encoding~\cite{instantngp} for the representation of temporal deformation significantly enhances the ability to reconstruct general videos. This formulation facilitates tracking the deformation of complex entities such as water and smog. However, the heightened capability of the deformation field presents a challenge in estimating a natural canonical image. An unnatural canonical image can also estimate the corresponding deformation field with a faithful reconstruction. To navigate this challenge, we suggest employing annealed hash during training. Initially, a smooth deformation grid is utilized to identify a coarse solution applicable to all rigid motions, with high-frequency details added gradually. Through this coarse-to-fine training, the representation achieves a balance between the naturalness of the canonical and the faithfulness of the reconstruction. We observe a noteworthy enhancement in reconstruction quality compared to preceding methods. This improvement is quantified as an approximately 4.4 increase in PSNR, along with an observable increase in the naturalness of the canonical image. Our optimization process requires a mere approximate 300 seconds to estimate the canonical image with the deformation field while the previous implicit layered representations\cite{kasten2021layered} takes more than 10 hours. 

Building upon our proposed content deformation field, we illustrate lifting 
image processing tasks such as prompt-guided image translation, super-resolution, and segmentation—to the more dynamic realm of video content. Our approach to prompt-guided video-to-video translation employs ControlNet~\cite{zhang2023adding} on the canonical image, propagating the translated content via the learned deformation. The translation process is conducted on a single canonical image and obviates the need for time-intensive inference models (\textit{e.g.}, Diffusion models) across all frames. Our translation outputs exhibit marked improvements in temporal consistency and texture quality over the state-of-the-art zero-shot video translations with generative models~\cite{wu2022tune,qi2023fatezero}. When contrasted with Text2Live, which relies on a neural layered atlas, our model is proficient in handling more complex motion, producing more natural canonical images, and thereby achieving superior translation results. Additionally, we extend the application of image algorithms such as super-resolution, semantic segmentation, and keypoints detection to the canonical image, leading to their practical applications in video contexts. This includes video super-resolution, video object segmentation, video keypoints tracking, among others. Our proposed representation consistently delivers superior temporal consistency and high-fidelity synthesized frames, demonstrating its potential as a groundbreaking tool in video processing.



%% file: sections/2.related.tex
\section{Related Work}\label{sec:related}

\noindent\textbf{Implicit Neural Representations.}
Implicit representations in conjunction with coordinate-based Multilayer Perceptrons (MLPs) have demonstrated its powerful capability in accurately representing images~\cite{tancik2020fourier, sitzmann2020implicit, chen2021learning}, videos~\cite{li2021neural, sitzmann2020implicit, kasten2021layered, ye2022deformable}, and 3D/4D representations~\cite{park2019deepsdf, michalkiewicz2019implicit, peng2020convolutional, mildenhall2020nerf, wang2021neus, park2021nerfies, park2021hypernerf}. These techniques have been employed in a range of applications, including novel view synthesis~\cite{mildenhall2020nerf}, image super-resolution~\cite{chen2021learning}, and 3D/4D Reconstruction~\cite{wang2021neus, wang2022neus2}.  Furthermore, for the purpose of speeding up the training, a various of acceleration~\cite{instantngp,yu_and_fridovichkeil2021plenoxels} techniques have been explored to replace the original Fourier positional encoding with some discrete representation like multi-resolution feature grid or hash table. Moreover, the adoption of an implicit deformation field~\cite{park2021nerfies, park2021hypernerf, pumarola2021d, li2022neural} has displayed a remarkable capability to overfit dynamic scenes.  Inspired by these works, our primary objective is to reconstruct videos by utilizing a canonical image which inherit semantics for video processing purposes. 

\begin{figure*}[t]
    \centering
    \includegraphics[width=1.0\linewidth]{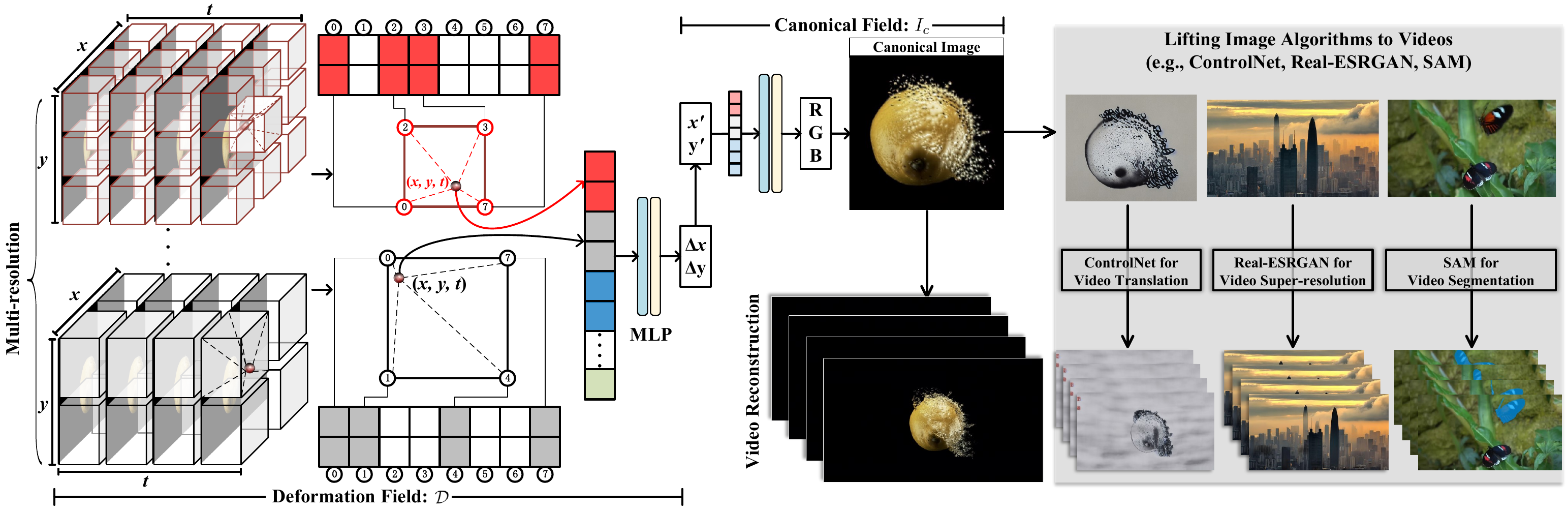}
    \vspace{-18pt}
    \caption{%
        \textbf{Illustration of the proposed video representation}, \method, which factorizes an arbitrary video into a 2D content canonical field and a 3D temporal deformation field.
        Each field is implemented with a multi-resolution 2D or 3D hash table using an efficient MLP.
        Such a new type of representation naturally supports \textit{lifting image algorithms for video processing}, in the way of directly applying the established algorithm on the canonical image (\textit{i.e.}, rendered from the canonical content field) and then propagating the results along the time axis through the temporal deformation field.
    }
    \label{fig:method}
    \vspace{-7pt}
\end{figure*}

\noindent\textbf{Consistent Video Editing.}
Our research is closely aligned with the domain of consistent video editing~\cite{jamrivska2019stylizing, lei2022deep, kasten2021layered, lu2020layered}, which predominantly features two primary approaches: propagation-based methods and layered representation-based techniques. Propagation-based methods~\cite{jampani2017video, jamrivska2019stylizing, texler2020interactive, jabri2020space, wang2019learning, ruder2016artistic} center on editing an initial frame and subsequently disseminating those edits throughout the video sequence. While this approach offers advantages in terms of computational efficiency and simplicity, it may be prone to inaccuracies and inconsistencies during the propagation of edits, particularly in situations characterized by complex motion or occlusion. Conversely, layered representation-based techniques~\cite{kasten2021layered, lu2020layered,  lu2022associating, rav2008unwrap, shade1998layered} entail decomposing a video into distinct layers, thereby facilitating greater control and flexibility during the editing process. 
Text2Live~\cite{bar2022text2live} introduces the application of CLIP~\cite{radford2021learning} models for video editing by modifying an optimized atlas~\cite{kasten2021layered} using text inputs, thereby yielding temporally consistent video editing results. Our work bears similarities to Text2Live in the context of employing an optimized representation for videos. However, our methodology diverges in several aspects: we optimize a more semantically-aware canonical representation incorporating a hash-based deformable design and attain higher-fidelity video processing.


\noindent\textbf{Video Processing via Generative Models.}
The advancement of diffusion models has markedly enhanced the synthesis quality of text-to-image generation~\cite{ho2020denoising, dhariwal2021diffusion, song2020denoising}, surpassing the performance of prior methodologies~\cite{mansimov2015generating, reed2016learning, xu2018attngan, zhang2017stackgan}. State-of-the-art diffusion models, such as GLIDE~\cite{nichol2021glide}, Dall-E 2~\cite{ramesh2021zero, ramesh2022hierarchical}, Stable Diffusion~\cite{rombach2022high}, and Imagen~\cite{saharia2022photorealistic}, have been trained on millions of images, resulting in exceptional generative capabilities. While existing text-to-image (T2I) models enable free-text generation, incorporating additional conditioning factors~\cite{ruiz2023dreambooth, mou2023t2i, tumanyan2023plug, wang2022pretraining, brooks2023instructpix2pix, hertz2022prompt, zhang2023adding, cao2023masactrl} such as edge, depth map, and normal map is essential for achieving precise control. In an effort to enhance controllability, researchers have proposed several approaches. For instance, PITI~\cite{wang2022pretraining} involves retraining an image encoder to map latents to the T2I latent space. InstructPix2Pix~\cite{brooks2023instructpix2pix}, on the other hand, fine-tunes T2I models using synthesized image condition pairs. ControlNet~\cite{zhang2023adding} introduces additional control conditions for Stable Diffusion through an auxiliary branch, thereby generating images that faithfully adhere to input condition maps.  A recent research direction concentrates on the processing of videos utilizing text-to-image (T2I) models exclusively. Approaches like Tune-A-Video~\cite{wu2022tune}, Text2Video-Zero~\cite{khachatryan2023text2video}, FateZero~\cite{qi2023fatezero}, Vid2Vid-Zero~\cite{wang2023zero}, and Video-P2P~\cite{liu2023video} explore the latent space of DDIM~\cite{song2020denoising} and incorporate cross-frame attention maps to facilitate consistent generation. Nevertheless, these methods may experience compromised temporal consistency due to the inherent randomness of generation, and the control condition may not be achieved with precision.


Text-to-video generation has emerged as a prominent research area in recent years, with prevalent approaches encompassing the training of diffusion models or autoregressive transformers on extensive datasets. Although text-to-video architectures such as NUWA~\cite{wu2022nuwa}, CogVideo~\cite{hong2022cogvideo}, Phenaki~\cite{villegas2022phenaki}, Make-A-Video~\cite{singer2022make}, Imagen Video~\cite{ho2022imagen}, and Gen-1~\cite{esser2023structure} are capable of generating video frames that semantically correspond to the input text, they may exhibit limitations in terms of precise control over video conditions or low resolution due to substantial computational demands.





%% file: sections/3.method.tex
\section{Method} \label{sec:method}

\noindent\textbf{Problem Formulation.}
Given a video $V$ comprised of frames $\{I_1, I_2,..., I_N\}$, one can naively apply the image processing algorithm $\mathcal{X}$ to each frame individually for corresponding video tasks, yet may observe undesirable inconsistencies across frames. An alternative strategy involves enhancing algorithm $\mathcal{X}$ with a temporal module, which requires additional training on video data. However, simply introducing a temporal module is hard to guarantee theoretical consistency and may result in performance degradation due to insufficient training data.

Motivated by these challenges, we propose representing a video $\mathcal{V}$ using a flattened canonical image $I_c$ and a deformation field $\mathcal{D}$. By applying the image algorithm $\mathcal{X}$ on $I_c$, we can effectively propagate the effect to the whole video with the learned deformation field. This novel video representation serves as a crucial bridge between image algorithms and video tasks, allowing directly lifting of state-of-the-art image methodologies to video applications.




The proposed representations ought to exhibit the following essential characteristics:

\begin{itemize}[leftmargin=10pt]
\vspace{-5pt}
\setlength{\itemsep}{0pt}
\setlength{\parsep}{0pt}
\setlength{\parskip}{0pt}
\setlength{\itemindent}{0pt}
\item \textbf{Fitting Capability for Faithful Video Reconstruction.} The representation should possess the ability to accurately fit large rigid or non-rigid deformations in videos.
\item \textbf{Semantic Correctness of the Canonical Image.} A distorted or semantically incorrect canonical image can lead to decreased image processing performance, especially considering that most of these processes are trained on natural image data.
\item \textbf{Smoothness of the Deformation Field.} The assurance of smoothness in the deformation field is an essential feature that guarantees temporal consistency and correct propagation.
\end{itemize}

\subsection{Content Deformation Fields}
\label{subsec:CoDeF}
Inspired by the dynamic NeRFs~\cite{park2021nerfies, park2021hypernerf}, we propose to represent the video in two distinct components: the canonical field and the deformation field.
These two components are realized through the employment of a 2D and a 3D hash table, respectively. To enhance the capacity of these hash tables, two minuscule MLPs are integrated.
We present our proposed representation for reconstructing and processing videos, as illustrated in~\cref{fig:method}. Given a video $\mathcal{V}$ comprising frames $\{I_1, I_2,...,I_N\}$, we train an implicit deformable model tailored to fit these frames. The model is composed of two coordinate-based MLPs: the deformation field $\mathcal{D}$ and the canonical field $\mathcal{C}$. 

The canonical field $\mathcal{C}$ serves as a continuous representation encompassing all flattened textures present in the video $\mathcal{V}$. It is defined by a function $\mathbf{F}: \mathbf{x} \rightarrow \mathbf{c}$, which maps a 2D position $\mathbf{x}:(x,y)$ to a color $\mathbf{c}:(r,g,b)$. In order to speed up the training and enable the network to capture the high-frequency details, we adopt the multi-resolution hash encoding $\gamma_{\text{2D}}: \mathbb{R}^2 \rightarrow \mathbb{R}^{2+F{\times}L}$ to map the coordinate $\mathbf{x}$ into a feature vector, where $L$ is the number of levels for multi-resolution and $F$ is the number of feature dimensions for per layer. The function $\gamma_{\text{2D}}(\mathbf{x})=(\mathbf{x},\mathbf{F}_1(\mathbf{x}), ..., \mathbf{F}_L(\mathbf{x}))$ facilitates the model's ability to capture high-frequency details, where $\mathbf{F}_i(\mathbf{x})$ is the features linearly interpolated by $\mathbf{x}$ at $i^{\text{th}}$ resolution.
The deformation field $\mathcal{D}$ captures the observation-to-canonical deformation for every frame within a video. For a specific frame $I_{i}$, $\mathcal{D}$ establishes the correspondence between the observed and canonical positions. Dynamic NeRFs~\cite{park2021nerfies,park2021hypernerf} implement the deformation field in 3D space by using the Fourier positional encoding and an extra learnable time code. This implementation ensures the smoothness of the deformation field. 
Nevertheless, this straightforward implementation can not be seamlessly transferred into video representation for two reasons (\textit{i.e.} low training efficiency and inadequate representative capability). 
Therefore, we propose to represent the deformation field as a 3D hash table with a tiny MLP following. 
Specifically, an arbitrary position $\mathbf{x}$ in the $t^{th}$ frame is first encoded by a 3D hash encoding function $\gamma_{\text{3D}}(\mathbf{x}, t)$ to get high-dimension features. Then a tiny MLP $\mathcal{D}: (\gamma_{\text{3D}}(\mathbf{x}, t)) \rightarrow \mathbf{x}'$ maps the embedded features its corresponding position $\mathbf{x}'$ in canonical field.
We elaborate our 3D hash encoding based deformation field in detail as follows.

\noindent\textbf{3D Hash Encoding for Deformation Field.}
Specifically, an arbitrary point in the video can be conceptualized as a position $\mathbf{x_{\text{3D}}}: (x,y,t)$ within an orthogonal 3D space.We represent our video space using the 3D hash encoding technique, as depicted on the left side of~\cref{fig:method}. This technique encapsulates the 3D space as a multi-resolution feature grid. The term \textit{multi-resolution} refers to a composition of grids with varying degrees of resolution, and \textit{feature grid} denotes a grid populated with learnable features at each vertex. In our framework, the multi-resolution feature grid is organized into $L$ distinct levels. The dimensionality of the learnable features is represented as $F$. Furthermore, the resolution of the $l^{\text{th}}$ layer, denoted as $N_{l}$, exhibits a geometric progression between the coarsest and finest resolutions, denoted collectively as $[N_{\text{min}},N_{\text{max}}]$, using
\begin{equation}
    N_{l}=\lfloor N_{\text{min}} \cdot b^{l} \rfloor, \\
    b=\text{exp}\left( \frac{\ln{N_{\text{max}}}-\ln{N_{\text{min}}}}{L-1} \right).
\end{equation}
Considering the queried points $\mathbf{x}_{\text{3D}}$ at $l^{\text{th}}$ layer, the input coordinate 
is scaled by that level’s grid resolution. And the queried features of $\mathbf{x}_{\text{3D}}$ are tri-linear interpolated from its 8-neighboring corner points(seen in \cref{fig:method}). For attaining the corner points of $\mathbf{x}_{\text{3D}}$, rounding down and up are first operated as
\begin{equation}
    \lfloor \mathbf{x}_{\text{3D}}^{l} \rfloor = \lfloor \mathbf{x}_{\text{3D}} \cdot N_{l} \rfloor, \lceil \mathbf{x}_{\text{3D}}^{l} \rceil = \lceil \mathbf{x}_{\text{3D}} \cdot N_{l} \rceil, 
\end{equation}
and we map its each corner to an entry in the level’s respective feature vector array,
which has fixed size of at most $T$. For the coarse level, the parameters of low resolution grid are fewer than $T$, where the mapping is $1:1$. Thus, the features can be directly looked up by its index. On the contrary. For the finer resolution, the point is mapped by the hash function, 
\begin{equation}
    h(\mathbf{x}_{\text{3D}}^{l})=\left( \oplus^{d}_{i=1} x_i \pi_{i}\right) \quad \text{mod}\quad T, 
\end{equation}
where $\oplus$ denotes the bit-wise XOR operation and $\{\pi_{i}\}$ are unique large prime numbers following ~\cite{instantngp}. 

The output color value at coordinate $\mathbf{x}$ for frame $t$ can be computed as
\begin{equation}
    \mathbf{c} = \mathcal{C}(\mathcal{D}(\gamma_{\text{3D}}(\mathbf{x}, t))).
\end{equation} This output can be supervised using the ground truth color present in the input frame.

\subsection{Model Design}
\label{subsec:model_design}

The proposed representation can effectively model and reconstruct both the canonical content and the temporal deformation for an arbitrary video. However, it faces challenges in meeting the requirements for robust video processing. In particular, while 3D hash deformation possesses powerful fitting capability, it compromises the smoothness of temporal deformation. This trade-off makes it notably difficult to maintain the inherent semantics of the canonical image, creating a significant barrier to the adaptation of established image algorithms for video use. To achieve precise video reconstruction while preserving the inherent semantics of the canonical image, we propose the use of annealed multi-resolution hash encoding. To further enhance the smoothness of deformation, we introduce flow-based consistency. In challenging cases, such as those involving large occlusions or complex multi-object scenarios, we suggest utilizing additional semantic information. This can be achieved by using semantic masks in conjunction with the grouped deformation fields.


\noindent\textbf{Annealed 3D Hash Encoding for Deformation.}
For the finer resolution, the hash encoding enhance the complex deformation fitting performance but introducing the discontinuity and distortion in canonical field (Seen in~\cref{fig:ablation_anneal}). Inspired by the annealed strategy utilized in dynamic NeRFs~\cite{park2021nerfies}, we employ the annealed hash encoding technique for progressive frequency filter for deformation. More specifically, we use a progressive controlling weights for those features interpolated in different resolution. The weight for the $l^{\text{th}}$ layer in training step $k$ is computed as
\begin{equation}
w_j(k) = \frac{1-\cos(\pi \cdot \text{clamp}(m(j-N_{\text{beg}})/N_{\text{step}}, 0, 1))}{2},
\end{equation}
where $N_{\text{beg}}$ is a predefined step for beginning annealing and $m$ represents a hyper parameters for controlling the annealing speed, and $N_{\text{step}}$ is the number for annealing step.

\noindent\textbf{Flow-guided Consistency Loss.} 
Corresponding points identified by flows with high confidence should be the same points in the canonical field.
We compute the flow-guided consistency loss according to this observation.
For two consecutive frames $I_i$ and $I_{i+1}$, we employ RAFT\cite{teed2020raft} to detect the forward flows $\mathcal{F}_{i \rightarrow i+1}$ and backward flows $\mathcal{F}_{i+1 \rightarrow i}$. The confident region of a frame $I_i$ can be defined as
\begin{equation}
     M_{\text{flow}} = |\text{Warp}(\text{Warp}(I_i, \mathcal{F}_{i \rightarrow i+1}), \mathcal{F}_{i+1 \rightarrow i}) - I_i| < \epsilon,
\end{equation}
where $\epsilon$ represents a hyperparameter for the error threshold.

This loss can be formulated as
%
\begin{equation}
\mathcal{L}_{\text{flow}} = \scalebox{0.7}{$\displaystyle\sum \|  \mathcal{D}(\gamma_{3D}(\mathbf{x}, t))  - \mathcal{D}(\gamma_{3D}(\mathbf{x} + \mathcal{F}_{t \rightarrow t+1}^{\mathbf{x}} , t+1)) -  \mathcal{F}_{t \rightarrow t+1}^{\mathbf{x}} \| * M_{\text{flow}}^{\mathbf{x}}$},
\end{equation}
where $\mathcal{F}_{t \rightarrow t+1}^{\mathbf{x}}$ and $M_{\text{flow}}^{\mathbf{x}}$ are the optical flow and the flow confidence at $\mathbf{x}$ . The flow loss efficiently regularize the smoothness of the deformation field especially for the smooth region.

\begin{figure}[t]
    \centering
    \includegraphics[width=1.0\linewidth]{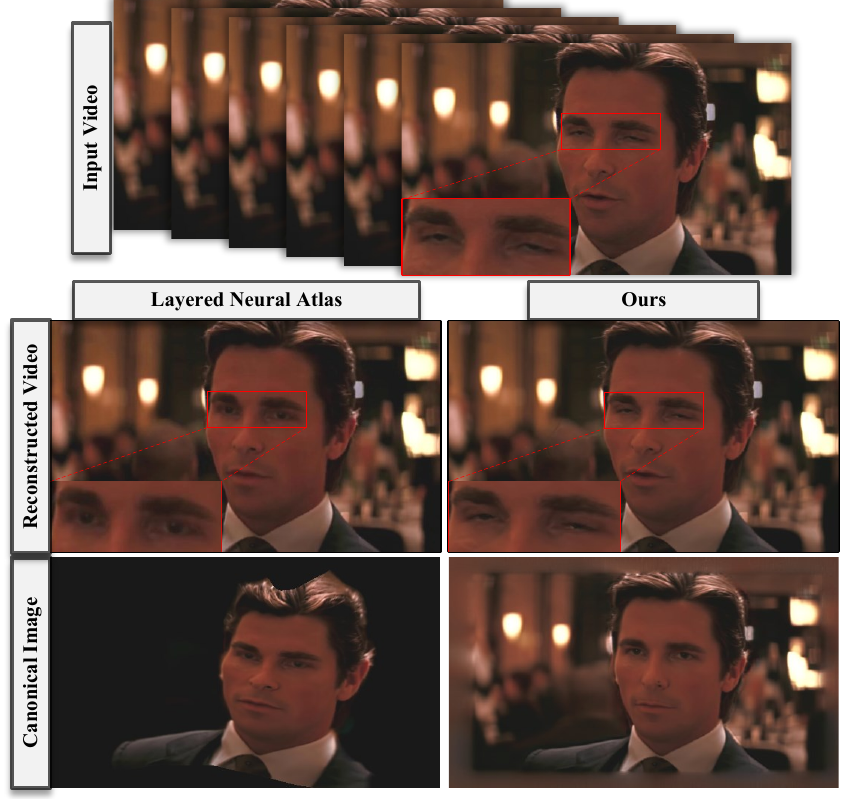}
    \vspace{-18pt}
    \caption{%
        \textbf{Qualitative comparison} between layered neural atlas~\cite{kasten2021layered} and our \method regarding \textit{video reconstruction}, which reflects the capacity of the video representation and also plays a fundamental role in faithful video processing. 
        Details are best appreciated when zoomed in. 
    }
    \label{fig:reconstruction_comparison}
    \vspace{-5pt}
\end{figure}

\begin{figure*}[t]
    \centering
    \includegraphics[width=1.0\linewidth]{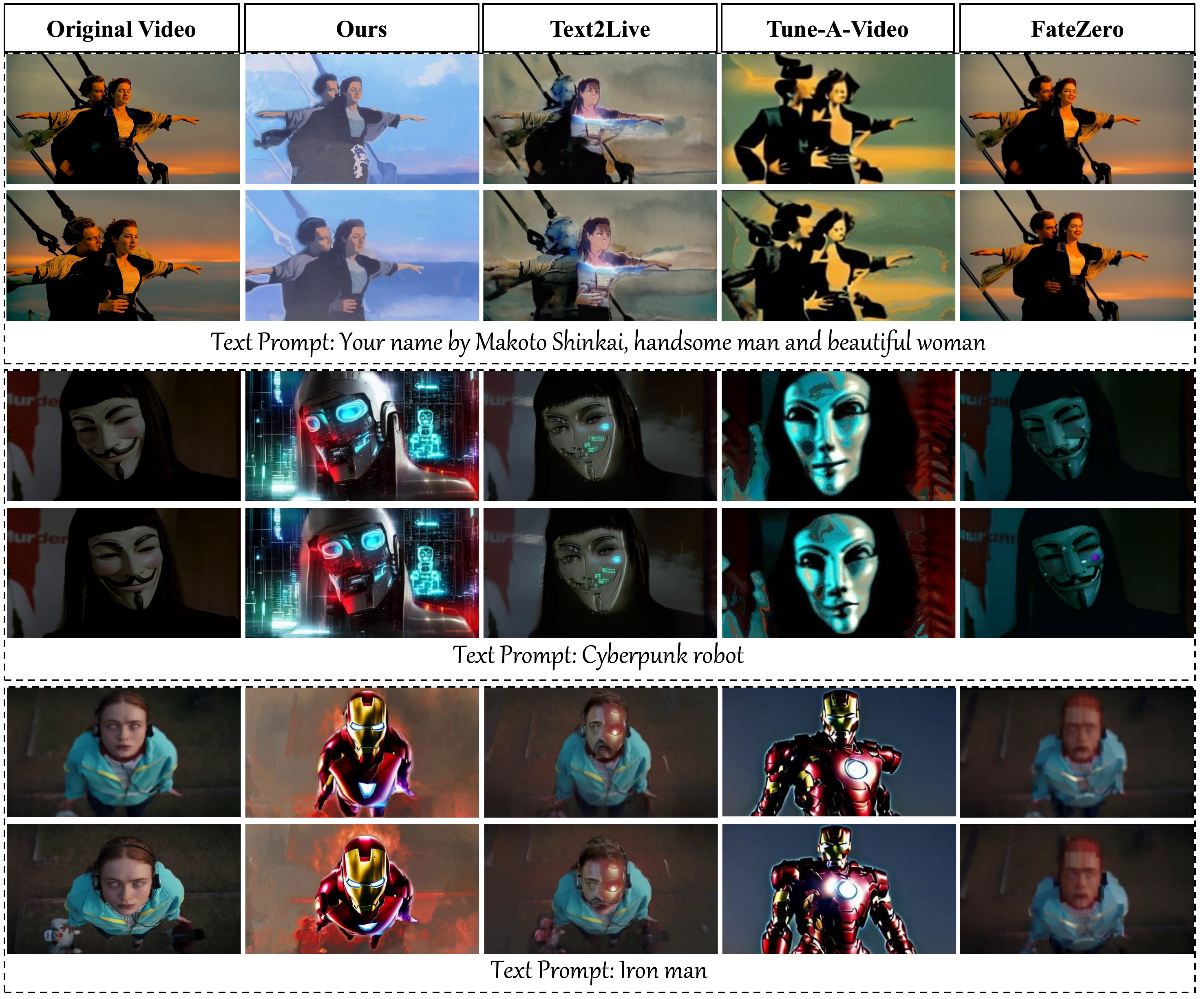}
    \vspace{-20pt}
    \caption{%
        \textbf{Qualitative comparison} on the task of \textit{text-guided video-to-video translation} across different methods, including Text2Live~\cite{bar2022text2live}, Tune-A-Video~\cite{wu2022tune}, FateZero~\cite{qi2023fatezero}, and directly lifting ControlNet~\cite{zhang2023adding} through our \method.
        We strongly encourage the readers to see the videos on the project page for a detailed evaluation of temporal consistency and synthesis quality.
    }
    \vspace{-5pt}
\label{fig:qualitative_comparison}  
\end{figure*}



\noindent\textbf{Grouped Content Deformation Fields.} Although the representation can learn to reconstruct a video using a single content deformation field, complex motions arising from overlapped multi-objects may lead to conflicts within one canonical. Consequently, the boundary region might suffer from inaccurate reconstruction. For challenging instances featuring large occlusions, we propose an option to introduce the layers corresponding to multiple content deformation fields. These layers would be defined based on semantic segmentation, thereby improving the accuracy and robustness of video reconstruction in these demanding scenarios. We leverage the Segment-Anything-track (SAM-track)~\cite{cheng2023segment} to attain the segmentation of each video frame $I_i$ into $K$ semantic layers with mask ${M_0^i,...,M_{K-1}^i}$. And for each layer, we use a group of canonical fields and deformation fields to represent those separate motion of different objects. These models are subsequently formulated as groups of implicit fields: $\mathcal{D}:\{\mathcal{D}_1,...,\mathcal{D}_K\}, \mathcal{C}:\{\mathcal{C}_1,...,\mathcal{C}_K\}$.  In theory, for semantic layer $k$ in frame $i$, it is sufficient to sample pixels in the region $M_k^i$ for efficient reconstruction.
However, hash encoding can result in random and unstructured patterns in unsupervised regions, which decreases the performance of image-based models trained on natural images.
To tackle this issue, we sample a number of points outside of the region $M_k^i$ and train them using $L_2$ loss with the ground truth color.
In this way, we effectively regularize ${\bar{M}_k^i}$ with the background loss  $\mathcal{L}_{\text{bg}}$. Consequently, the canonical image attains a more natural appearance, leading to enhanced processing results.

\noindent\textbf{Training Objectives.}
The representation is trained by minimizing the objective function $\mathcal{L}_{\text{rec}}$. This function corresponds to the $L_2$ loss between the ground truth color and the predicted color $\mathbf{c}$ for a given coordinate $\mathbf{x}$. To regularize and stabilize the training process, we introduce additional regularization terms as previously discussed. The total loss is calculated using the following equation
\begin{equation}
\mathcal{L} = \mathcal{L}_{\text{rec}} +\lambda_1 * \mathcal{L}_{\text{flow}},
\end{equation}
where $\lambda_1$ represents the hyper-parameters for loss weights. It's important to note that when training the grouped deformation field, we include an additional regularizer, denoted as $\lambda_2 * \mathcal{L}_{\text{bg}}$.

\begin{figure*}[t]
    \centering
    \includegraphics[width=1.0\linewidth]{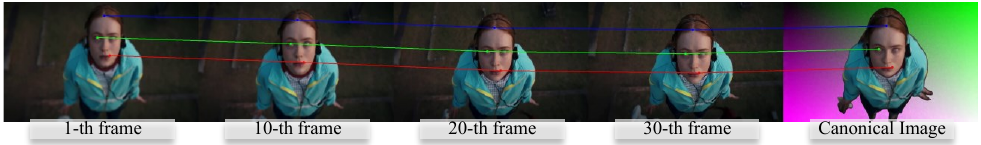}
    \vspace{-18pt}
    \caption{%
        \textbf{Visualization of point correspondence across frames}, which is directly extracted from the temporal deformation field after reconstructing the video with \method.
        %
    }
    \label{fig:tracking}  
    \vspace{-5pt}
\end{figure*}

\subsection{Application to Consistent Video Processing} 
Upon the optimization of the content deformation field, the canonical image $I_c$ is retrieved by setting the deformation of all points to zero. It is important to note that the size of the canonical image can be flexibly adjusted to be larger than the original image size depending on the scene movement observed in the video, thereby allowing more content to be included. The canonical image $I_c$ is then utilized in executing various downstream algorithms for consistent video processing. We evaluated the following state-of-the-art (SOTA) algorithms: \textit{(1) ControlNet}~\cite{zhang2023adding}: Used for prompt-guided video-to-video translation.
\textit{(2) Segment-anything (SAM)}~\cite{kirillov2023segment}: Applied for video object tracking.
\textit{(3) R-ESRGAN}~\cite{wang2021real}: Employed for video super-resolution. Additionally, the canonical image allows users to conveniently edit the video by directly modifying the image. We further illustrate this capability through multiple manual video editing examples.

%% file: sections/4.exp.tex
\section{Experiments}\label{sec:exp}

\subsection{Experimental Setup}

We conduct experiments to underscore the robustness and versatility of our proposed method.
Our representation is robust with a variety of deformations, encompassing rigid and non-rigid objects, as well as complex scenarios such as smog.
The default parameters for our experiments are set with the anneal begin and end steps at 4000 and 8000, respectively.
The total iteration step is capped at 10000. 
On a single NVIDIA A6000 GPU, the average training duration is approximately 5 minutes when utilizing 100 video frames.
%
%
It should be noted that the training time varies with several factors such as the length of the video, the type of motion, and the number of layers.
By adjusting the training parameters accordingly, the optimization duration can be varied from 1 to 10 minutes.

\begin{figure}[t]
    \centering
    \includegraphics[width=1.0\linewidth]{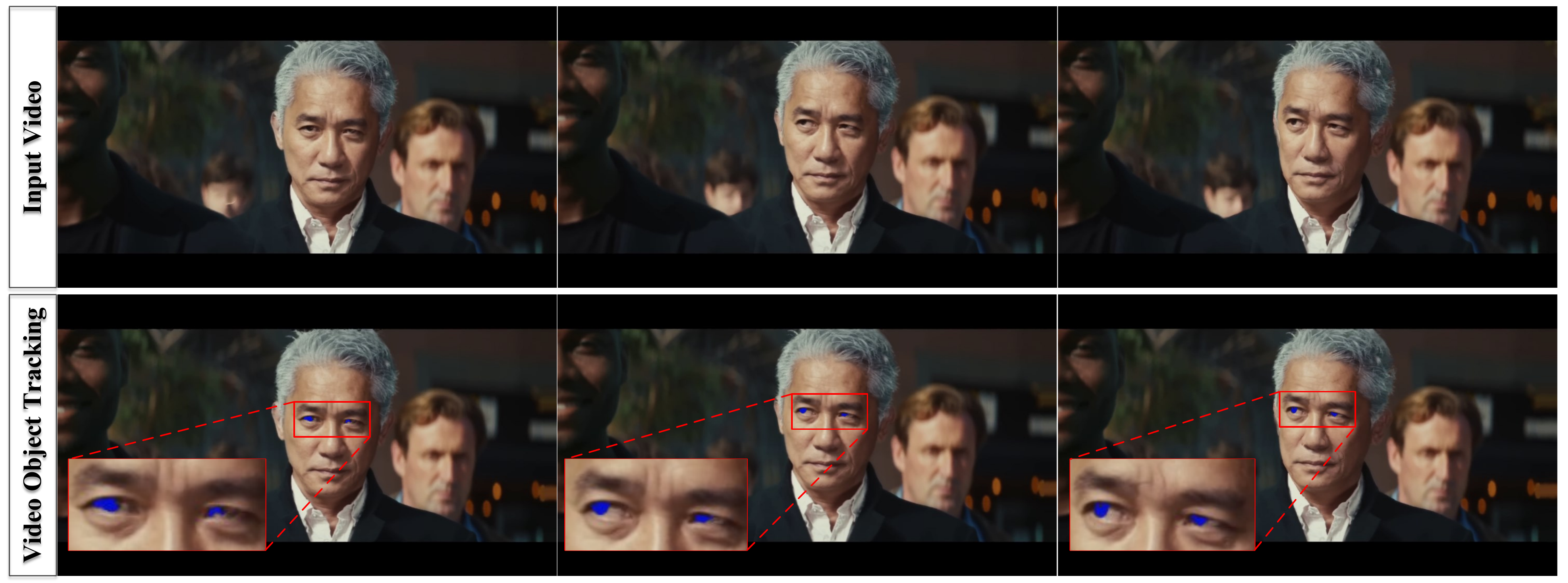}
    \vspace{-20pt}
    \caption{%
        \textbf{Video object tracking} results achieved by \textit{lifting} an image segmentation algorithm~\cite{kirillov2023segment} through our \method.
        %
    }
    \label{fig:object_track}  
    \vspace{-10pt}
\end{figure}

\begin{figure}[t]
    \centering
    \includegraphics[width=1.0\linewidth]{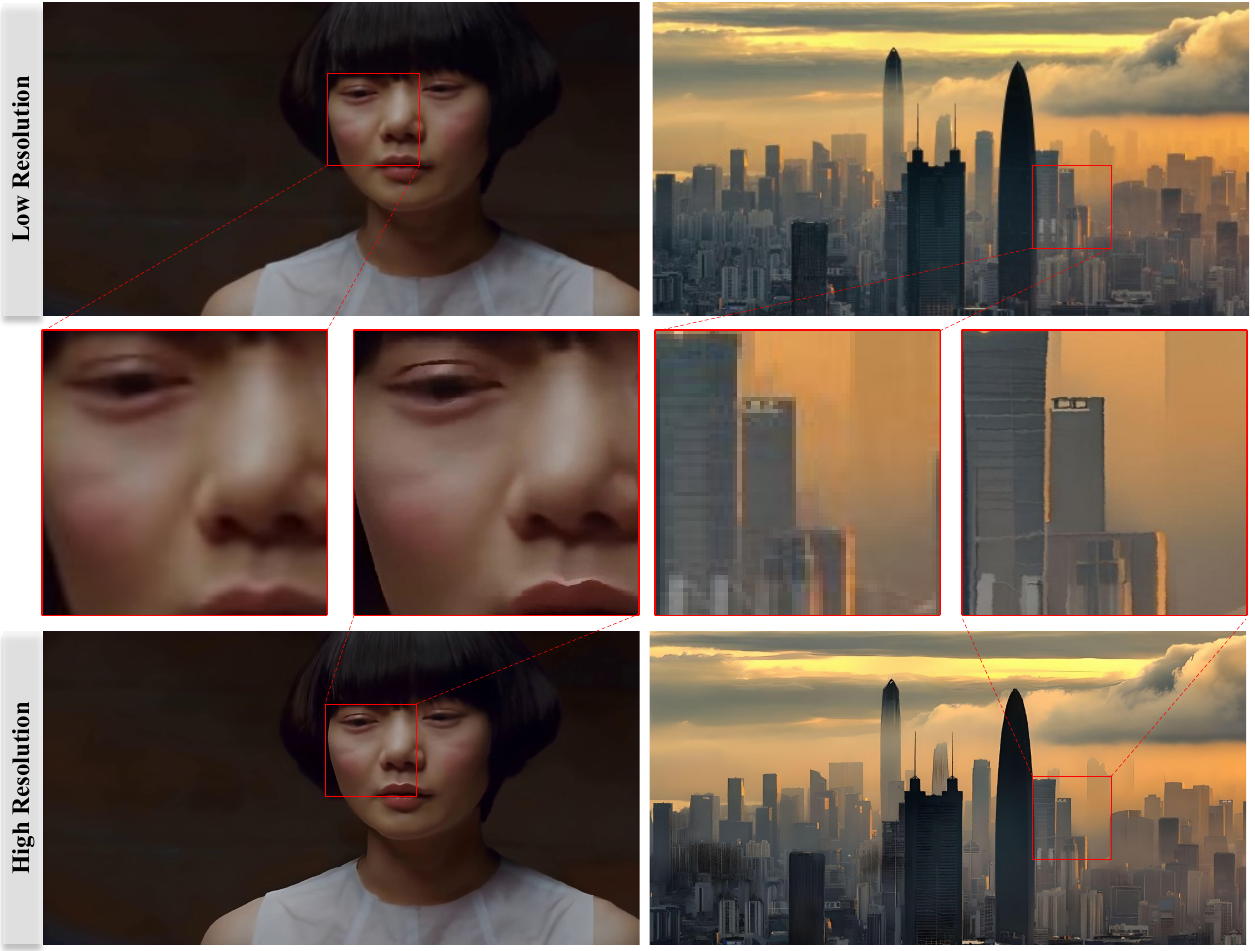}
    \vspace{-20pt}
    \caption{%
        \textbf{Video super-resolution} results achieved by \textit{lifting} an image super-resolution algorithm~\cite{wang2021real} through our \method.
        %
    }
    \label{fig:SR}  
    \vspace{-5pt}
\end{figure}

\subsection{Evaluation}
The evaluation of our representation is concentrated on two main aspects: the quality of the reconstructed video with the estimated canonical image, and the quality of downstream video processing. Owing to the lack of accurate evaluation metrics, conducting a precise quantitative analysis remains challenging. Nevertheless, we include a selection of quantitative results for further examination.

\noindent\textbf{Reconstruction Quality.}
 In a comparative analysis with the Neural Image Atlas, our model, as demonstrated in~\cref{fig:reconstruction_comparison}, exhibits superior robustness to non-rigid motion, effectively reconstructing subtle movements with heightened precision (\textit{e.g.} eyes blinking, face textures). Quantitatively, the video reconstruction PSNR of our algorithm on the collected video datasets is 4.4 dB higher. In comparison between the atlas and our canonical image, our results provide a more natural representation, and thus, facilitate the easier application of established image algorithms. Besides, our method makes a significant progress in training efficiency, ~\textit{i.e.}, 5 minutes (ours) \textit{vs.} 10 hours (atlas).
 

\noindent\textbf{Downstream Video Processing.}
We provide an expanded range of potential applications associated with the proposed representations, including video-to-video translation, video keypoint tracking, video object tracking, video super-resolution, and user-interactive video editing.

\noindent\textit{(a) Video-to-video Translation.}
By applying image translation to the canonical image, we can perform video-to-video translation. A qualitative comparison is presented encompassing several baseline methods that fall into three distinct categories: (1) per-frame inference with image translation models, such as ControlNet~\cite{zhang2023adding}; (2) layered video editing, exemplified by Text-to-live~\cite{bar2022text2live}; and (3) diffusion-based video translation, including Tune-A-Video~\cite{wu2022tune} and FateZero~\cite{qi2023fatezero}. As depicted in~\cref{fig:qualitative_comparison}, the per-frame image translation models yield high-fidelity content, accompanied by significant flickering. The alternative baselines exhibit compromised generation quality or comparatively low temporal consistency. The proposed pipeline effectively lifts image translation to video, maintaining the high quality associated with image translation algorithms while ensuring substantial temporal consistency. A thorough comparison is better appreciated by viewing the accompanying videos.

\noindent\textit{(b) Video Keypoint Tracking.}
By estimating the deformation field for each individual frame, it is feasible to query the position of a specific keypoint in one frame within the canonical space and subsequently identify the corresponding points present in all frames as in~\cref{fig:tracking}. We show the demonstration of tracking points in non-rigid objects such as fluids in the videos on the project page.

\noindent\textit{(c) Video Object Tracking.}
Using the segmentation algorithms on the canonical image, we are able to facilitate the propagation of masks throughout all video sequences leveraging the content deformation field. As illustrated in~\cref{fig:object_track}, our pipeline proficiently yields masks that maintain consistency across all frames.

\noindent\textit{(d) Video Super-resolution.}
By directly applying the image super-resolution algorithm to the canonical image, we can execute video super-resolution to generate high-quality video as in~\cref{fig:SR}. Given that the deformation is represented by a continuous field, the application of super-resolution will not result in flickering.

\begin{figure}[t]
    \centering
    \includegraphics[width=1.0\linewidth]{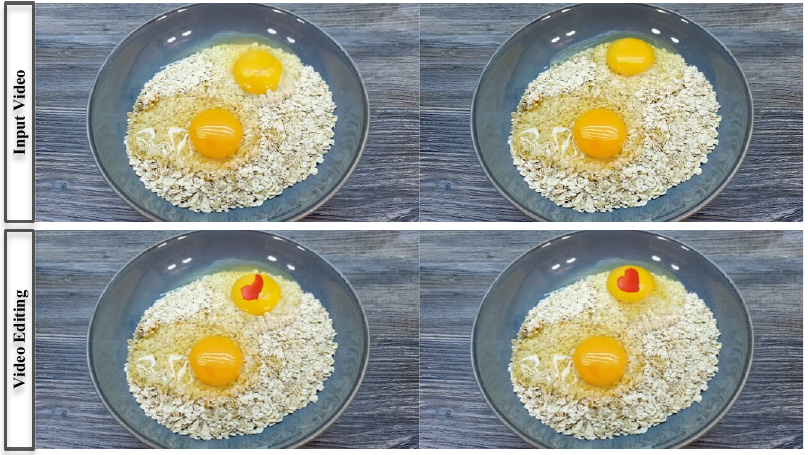}
    \vspace{-20pt}
    \caption{%
        \textbf{User interactive video editing} achieved by editing \textit{only one} image and propagating the outcomes along the time axis using our \method.
        %
        %
        We strongly encourage the readers to see the videos on the project page to appreciate the temporal consistency.
    }
    \label{fig:interactive_editing}  
    \vspace{-10pt}
\end{figure}

\noindent\textit{(e) User interactive Video Editing.}
Our representation allows for user editing on objects with unique styles without influencing other parts of the image. As exemplified in~\cref{fig:interactive_editing}, users can manually adjust content on the canonical image to perform precise edits in areas where the automatic editing algorithm may not be achieving optimal results.

\subsection{Ablation Study}

\begin{figure}[t]
    \centering
    \includegraphics[width=1.0\linewidth]{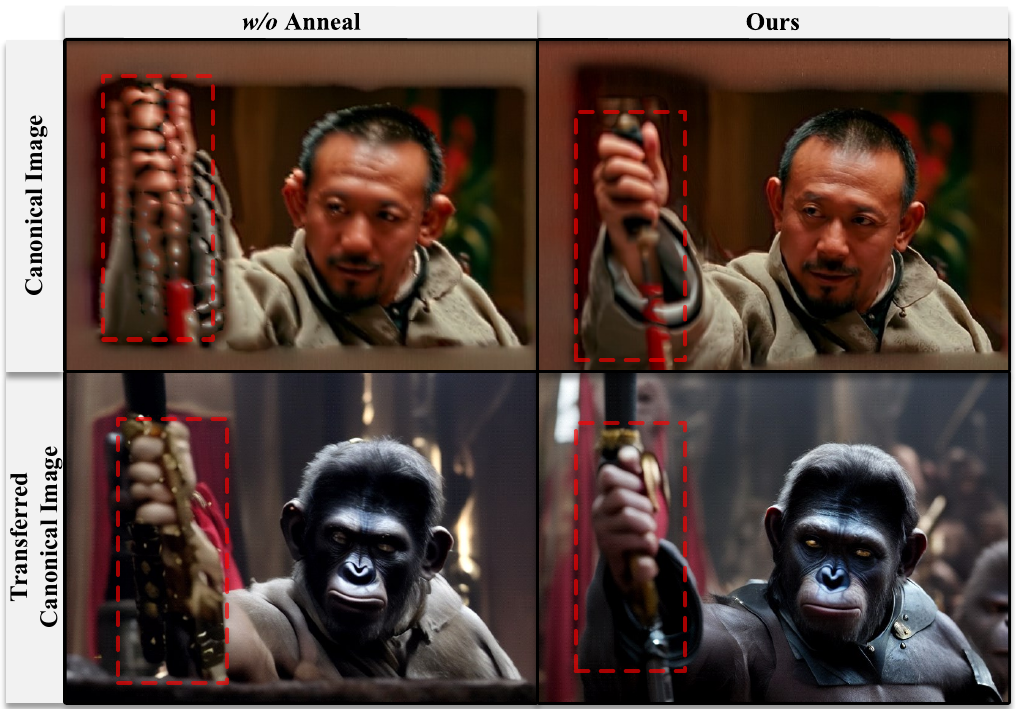}
    \vspace{-20pt}
    \caption{%
        \textbf{Ablation study} on the effectiveness of annealed hash.
        The unnaturalness in the canonical image will harm the performance of downstream tasks. 
    }
    \vspace{-5pt}
    \label{fig:ablation_anneal}  
\end{figure}

To validate the effect of the proposed modules, we conducted an ablation study. On substituting the 3D hash encoding with positional encoding, there is a notable decrease in the reconstruction PSNR of the video by 3.1 dB. In the absence of the annealed hash, the canonical image loses its natural appearance, as evidenced by the presence of multiple hands in~\cref{fig:ablation_anneal}. Furthermore, without incorporating the flow loss, smooth areas are noticeably affected by pronounced flickering. For a more extensive comparison, please refer to the videos on the project page.

%% file: sections/5.conclusion.tex
\section{Conclusion and Discussion}\label{sec:conclusion}

In this paper, we have investigated representing videos as content deformation fields, focusing on achieving temporally consistent video processing. Our approach demonstrates promising results in terms of both fidelity and temporal consistency. However, there remain several challenges to be addressed in future work.

One of the primary issues pertains to the per-scene optimization required in our methodology. We anticipate that advancements in feed-forward implicit field techniques~\cite{wang2021ibrnet, yu2021pixelnerf} could potentially be adapted to this direction. Another challenge arises in scenarios involving extreme changes in viewing points. To tackle this issue, the incorporation of 3D prior knowledge~\cite{haque2023instruct} may prove beneficial, as it can provide additional information and constraints. Lastly, the handling of large non-rigid deformations remains a concern. To address this, one potential solution involves employing multiple canonical images~\cite{park2021hypernerf}, which can better capture and represent complex deformations.

%% file: sections/6.ref.tex
{\small
\bibliographystyle{ieee_fullname}
\bibliography{ref}
}

%% file: main.bbl
\begin{thebibliography}{10}\itemsep=-1pt

\bibitem{bar2022text2live}
Omer Bar-Tal, Dolev Ofri-Amar, Rafail Fridman, Yoni Kasten, and Tali Dekel.
\newblock Text2live: Text-driven layered image and video editing.
\newblock In {\em Eur. Conf. Comput. Vis.}, 2022.

\bibitem{brooks2023instructpix2pix}
Tim Brooks, Aleksander Holynski, and Alexei~A Efros.
\newblock Instructpix2pix: Learning to follow image editing instructions.
\newblock In {\em IEEE Conf. Comput. Vis. Pattern Recog.}, 2023.

\bibitem{cao2023masactrl}
Mingdeng Cao, Xintao Wang, Zhongang Qi, Ying Shan, Xiaohu Qie, and Yinqiang
  Zheng.
\newblock Masactrl: Tuning-free mutual self-attention control for consistent
  image synthesis and editing.
\newblock {\em arXiv preprint arXiv:2304.08465}, 2023.

\bibitem{chen2021learning}
Yinbo Chen, Sifei Liu, and Xiaolong Wang.
\newblock Learning continuous image representation with local implicit image
  function.
\newblock In {\em IEEE Conf. Comput. Vis. Pattern Recog.}, 2021.

\bibitem{cheng2023segment}
Yangming Cheng, Liulei Li, Yuanyou Xu, Xiaodi Li, Zongxin Yang, Wenguan Wang,
  and Yi Yang.
\newblock Segment and track anything.
\newblock {\em arXiv preprint arXiv:2305.06558}, 2023.

\bibitem{dhariwal2021diffusion}
Prafulla Dhariwal and Alexander Nichol.
\newblock Diffusion models beat gans on image synthesis.
\newblock In {\em Adv. Neural Inform. Process. Syst.}, 2021.

\bibitem{esser2023structure}
Patrick Esser, Johnathan Chiu, Parmida Atighehchian, Jonathan Granskog, and
  Anastasis Germanidis.
\newblock Structure and content-guided video synthesis with diffusion models.
\newblock {\em arXiv preprint arXiv:2302.03011}, 2023.

\bibitem{haque2023instruct}
Ayaan Haque, Matthew Tancik, Alexei~A Efros, Aleksander Holynski, and Angjoo
  Kanazawa.
\newblock Instruct-nerf2nerf: Editing 3d scenes with instructions.
\newblock {\em arXiv preprint arXiv:2303.12789}, 2023.

\bibitem{hertz2022prompt}
Amir Hertz, Ron Mokady, Jay Tenenbaum, Kfir Aberman, Yael Pritch, and Daniel
  Cohen-Or.
\newblock Prompt-to-prompt image editing with cross attention control.
\newblock {\em arXiv preprint arXiv:2208.01626}, 2022.

\bibitem{ho2022imagen}
Jonathan Ho, William Chan, Chitwan Saharia, Jay Whang, Ruiqi Gao, Alexey
  Gritsenko, Diederik~P Kingma, Ben Poole, Mohammad Norouzi, David~J Fleet,
  et~al.
\newblock Imagen video: High definition video generation with diffusion models.
\newblock {\em arXiv preprint arXiv:2210.02303}, 2022.

\bibitem{ho2020denoising}
Jonathan Ho, Ajay Jain, and Pieter Abbeel.
\newblock Denoising diffusion probabilistic models.
\newblock In {\em Adv. Neural Inform. Process. Syst.}, 2020.

\bibitem{hong2022cogvideo}
Wenyi Hong, Ming Ding, Wendi Zheng, Xinghan Liu, and Jie Tang.
\newblock Cogvideo: Large-scale pretraining for text-to-video generation via
  transformers.
\newblock {\em arXiv preprint arXiv:2205.15868}, 2022.

\bibitem{jabri2020space}
Allan Jabri, Andrew Owens, and Alexei Efros.
\newblock Space-time correspondence as a contrastive random walk.
\newblock In {\em Adv. Neural Inform. Process. Syst.}, 2020.

\bibitem{jampani2017video}
Varun Jampani, Raghudeep Gadde, and Peter~V Gehler.
\newblock Video propagation networks.
\newblock In {\em IEEE Conf. Comput. Vis. Pattern Recog.}, 2017.

\bibitem{jamrivska2019stylizing}
Ond{\v{r}}ej Jamri{\v{s}}ka, {\v{S}}{\'a}rka Sochorov{\'a}, Ond{\v{r}}ej
  Texler, Michal Luk{\'a}{\v{c}}, Jakub Fi{\v{s}}er, Jingwan Lu, Eli Shechtman,
  and Daniel S{\`y}kora.
\newblock Stylizing video by example.
\newblock {\em ACM Trans. Graph.}, 38(4):1--11, 2019.

\bibitem{kasten2021layered}
Yoni Kasten, Dolev Ofri, Oliver Wang, and Tali Dekel.
\newblock Layered neural atlases for consistent video editing.
\newblock {\em ACM Trans. Graph.}, 40(6):1--12, 2021.

\bibitem{khachatryan2023text2video}
Levon Khachatryan, Andranik Movsisyan, Vahram Tadevosyan, Roberto Henschel,
  Zhangyang Wang, Shant Navasardyan, and Humphrey Shi.
\newblock Text2video-zero: Text-to-image diffusion models are zero-shot video
  generators.
\newblock {\em arXiv preprint arXiv:2303.13439}, 2023.

\bibitem{kirillov2023segment}
Alexander Kirillov, Eric Mintun, Nikhila Ravi, Hanzi Mao, Chloe Rolland, Laura
  Gustafson, Tete Xiao, Spencer Whitehead, Alexander~C Berg, Wan-Yen Lo, et~al.
\newblock Segment anything.
\newblock {\em arXiv preprint arXiv:2304.02643}, 2023.

\bibitem{lei2022deep}
Chenyang Lei, Yazhou Xing, Hao Ouyang, and Qifeng Chen.
\newblock Deep video prior for video consistency and propagation.
\newblock {\em IEEE Trans. Pattern Anal. Mach. Intell.}, 45(1):356--371, 2022.

\bibitem{li2022neural}
Tianye Li, Mira Slavcheva, Michael Zollhoefer, Simon Green, Christoph Lassner,
  Changil Kim, Tanner Schmidt, Steven Lovegrove, Michael Goesele, Richard
  Newcombe, et~al.
\newblock Neural 3d video synthesis from multi-view video.
\newblock In {\em IEEE Conf. Comput. Vis. Pattern Recog.}, 2022.

\bibitem{li2021neural}
Zhengqi Li, Simon Niklaus, Noah Snavely, and Oliver Wang.
\newblock Neural scene flow fields for space-time view synthesis of dynamic
  scenes.
\newblock In {\em IEEE Conf. Comput. Vis. Pattern Recog.}, 2021.

\bibitem{liu2023video}
Shaoteng Liu, Yuechen Zhang, Wenbo Li, Zhe Lin, and Jiaya Jia.
\newblock Video-p2p: Video editing with cross-attention control.
\newblock {\em arXiv preprint arXiv:2303.04761}, 2023.

\bibitem{lu2020layered}
Erika Lu, Forrester Cole, Tali Dekel, Weidi Xie, Andrew Zisserman, David
  Salesin, William~T Freeman, and Michael Rubinstein.
\newblock Layered neural rendering for retiming people in video.
\newblock {\em arXiv preprint arXiv:2009.07833}, 2020.

\bibitem{lu2022associating}
Erika Lu, Forrester Cole, Weidi Xie, Tali Dekel, Bill Freeman, Andrew
  Zisserman, and Michael Rubinstein.
\newblock Associating objects and their effects in video through coordination
  games.
\newblock In {\em Adv. Neural Inform. Process. Syst.}, 2022.

\bibitem{mansimov2015generating}
Elman Mansimov, Emilio Parisotto, Jimmy~Lei Ba, and Ruslan Salakhutdinov.
\newblock Generating images from captions with attention.
\newblock {\em arXiv preprint arXiv:1511.02793}, 2015.

\bibitem{michalkiewicz2019implicit}
Mateusz Michalkiewicz, Jhony~K Pontes, Dominic Jack, Mahsa Baktashmotlagh, and
  Anders Eriksson.
\newblock Implicit surface representations as layers in neural networks.
\newblock In {\em Int. Conf. Comput. Vis.}, 2019.

\bibitem{mildenhall2020nerf}
Ben Mildenhall, Pratul~P Srinivasan, Matthew Tancik, Jonathan~T Barron, Ravi
  Ramamoorthi, and Ren Ng.
\newblock Nerf: Representing scenes as neural radiance fields for view
  synthesis.
\newblock In {\em Eur. Conf. Comput. Vis.}, 2020.

\bibitem{mou2023t2i}
Chong Mou, Xintao Wang, Liangbin Xie, Jian Zhang, Zhongang Qi, Ying Shan, and
  Xiaohu Qie.
\newblock T2i-adapter: Learning adapters to dig out more controllable ability
  for text-to-image diffusion models.
\newblock {\em arXiv preprint arXiv:2302.08453}, 2023.

\bibitem{instantngp}
Thomas M{\"{u}}ller, Alex Evans, Christoph Schied, and Alexander Keller.
\newblock Instant neural graphics primitives with a multiresolution hash
  encoding.
\newblock {\em ACM Trans. Graph.}, 41(4):102:1--102:15, 2022.

\bibitem{nichol2021glide}
Alex Nichol, Prafulla Dhariwal, Aditya Ramesh, Pranav Shyam, Pamela Mishkin,
  Bob McGrew, Ilya Sutskever, and Mark Chen.
\newblock Glide: Towards photorealistic image generation and editing with
  text-guided diffusion models.
\newblock {\em arXiv preprint arXiv:2112.10741}, 2021.

\bibitem{park2019deepsdf}
Jeong~Joon Park, Peter Florence, Julian Straub, Richard Newcombe, and Steven
  Lovegrove.
\newblock Deepsdf: Learning continuous signed distance functions for shape
  representation.
\newblock In {\em IEEE Conf. Comput. Vis. Pattern Recog.}, 2019.

\bibitem{park2021nerfies}
Keunhong Park, Utkarsh Sinha, Jonathan~T Barron, Sofien Bouaziz, Dan~B Goldman,
  Steven~M Seitz, and Ricardo Martin-Brualla.
\newblock Nerfies: Deformable neural radiance fields.
\newblock In {\em Int. Conf. Comput. Vis.}, 2021.

\bibitem{park2021hypernerf}
Keunhong Park, Utkarsh Sinha, Peter Hedman, Jonathan~T. Barron, Sofien Bouaziz,
  Dan~B Goldman, Ricardo Martin-Brualla, and Steven~M. Seitz.
\newblock Hypernerf: A higher-dimensional representation for topologically
  varying neural radiance fields.
\newblock {\em ACM Trans. Graph.}, 40(6), 2021.

\bibitem{peng2020convolutional}
Songyou Peng, Michael Niemeyer, Lars Mescheder, Marc Pollefeys, and Andreas
  Geiger.
\newblock Convolutional occupancy networks.
\newblock In {\em Eur. Conf. Comput. Vis.}, 2020.

\bibitem{pumarola2021d}
Albert Pumarola, Enric Corona, Gerard Pons-Moll, and Francesc Moreno-Noguer.
\newblock D-nerf: Neural radiance fields for dynamic scenes.
\newblock In {\em IEEE Conf. Comput. Vis. Pattern Recog.}, 2021.

\bibitem{qi2023fatezero}
Chenyang Qi, Xiaodong Cun, Yong Zhang, Chenyang Lei, Xintao Wang, Ying Shan,
  and Qifeng Chen.
\newblock Fatezero: Fusing attentions for zero-shot text-based video editing.
\newblock {\em arXiv preprint arXiv:2303.09535}, 2023.

\bibitem{radford2021learning}
Alec Radford, Jong~Wook Kim, Chris Hallacy, Aditya Ramesh, Gabriel Goh,
  Sandhini Agarwal, Girish Sastry, Amanda Askell, Pamela Mishkin, Jack Clark,
  et~al.
\newblock Learning transferable visual models from natural language
  supervision.
\newblock In {\em Int. Conf. Mach. Learn.}, 2021.

\bibitem{ramesh2022hierarchical}
Aditya Ramesh, Prafulla Dhariwal, Alex Nichol, Casey Chu, and Mark Chen.
\newblock Hierarchical text-conditional image generation with clip latents.
\newblock {\em arXiv preprint arXiv:2204.06125}, 2022.

\bibitem{ramesh2021zero}
Aditya Ramesh, Mikhail Pavlov, Gabriel Goh, Scott Gray, Chelsea Voss, Alec
  Radford, Mark Chen, and Ilya Sutskever.
\newblock Zero-shot text-to-image generation.
\newblock In {\em Int. Conf. Mach. Learn.}, 2021.

\bibitem{rav2008unwrap}
Alex Rav-Acha, Pushmeet Kohli, Carsten Rother, and Andrew Fitzgibbon.
\newblock Unwrap mosaics: A new representation for video editing.
\newblock In {\em SIGGRAPH}, pages 1--11, 2008.

\bibitem{reed2016learning}
Scott~E Reed, Zeynep Akata, Santosh Mohan, Samuel Tenka, Bernt Schiele, and
  Honglak Lee.
\newblock Learning what and where to draw.
\newblock In {\em Adv. Neural Inform. Process. Syst.}, 2016.

\bibitem{rombach2022high}
Robin Rombach, Andreas Blattmann, Dominik Lorenz, Patrick Esser, and Bj{\"o}rn
  Ommer.
\newblock High-resolution image synthesis with latent diffusion models.
\newblock In {\em IEEE Conf. Comput. Vis. Pattern Recog.}, 2022.

\bibitem{ruder2016artistic}
Manuel Ruder, Alexey Dosovitskiy, and Thomas Brox.
\newblock Artistic style transfer for videos.
\newblock In {\em Pattern Recognition: 38th German Conference, GCPR 2016,
  Hannover, Germany, September 12-15, 2016, Proceedings 38}, pages 26--36.
  Springer, 2016.

\bibitem{ruiz2023dreambooth}
Nataniel Ruiz, Yuanzhen Li, Varun Jampani, Yael Pritch, Michael Rubinstein, and
  Kfir Aberman.
\newblock Dreambooth: Fine tuning text-to-image diffusion models for
  subject-driven generation.
\newblock In {\em IEEE Conf. Comput. Vis. Pattern Recog.}, 2023.

\bibitem{saharia2022photorealistic}
Chitwan Saharia, William Chan, Saurabh Saxena, Lala Li, Jay Whang, Emily~L
  Denton, Kamyar Ghasemipour, Raphael Gontijo~Lopes, Burcu Karagol~Ayan, Tim
  Salimans, et~al.
\newblock Photorealistic text-to-image diffusion models with deep language
  understanding.
\newblock In {\em Adv. Neural Inform. Process. Syst.}, 2022.

\bibitem{yu_and_fridovichkeil2021plenoxels}
{Sara Fridovich-Keil and Alex Yu}, Matthew Tancik, Qinhong Chen, Benjamin
  Recht, and Angjoo Kanazawa.
\newblock Plenoxels: Radiance fields without neural networks.
\newblock In {\em IEEE Conf. Comput. Vis. Pattern Recog.}, 2022.

\bibitem{shade1998layered}
Jonathan Shade, Steven Gortler, Li-wei He, and Richard Szeliski.
\newblock Layered depth images.
\newblock In {\em Proceedings of the 25th annual conference on Computer
  graphics and interactive techniques}, 1998.

\bibitem{singer2022make}
Uriel Singer, Adam Polyak, Thomas Hayes, Xi Yin, Jie An, Songyang Zhang, Qiyuan
  Hu, Harry Yang, Oron Ashual, Oran Gafni, et~al.
\newblock Make-a-video: Text-to-video generation without text-video data.
\newblock {\em arXiv preprint arXiv:2209.14792}, 2022.

\bibitem{sitzmann2020implicit}
Vincent Sitzmann, Julien Martel, Alexander Bergman, David Lindell, and Gordon
  Wetzstein.
\newblock Implicit neural representations with periodic activation functions.
\newblock In {\em Adv. Neural Inform. Process. Syst.}, 2020.

\bibitem{song2020denoising}
Jiaming Song, Chenlin Meng, and Stefano Ermon.
\newblock Denoising diffusion implicit models.
\newblock {\em arXiv preprint arXiv:2010.02502}, 2020.

\bibitem{tancik2020fourier}
Matthew Tancik, Pratul Srinivasan, Ben Mildenhall, Sara Fridovich-Keil, Nithin
  Raghavan, Utkarsh Singhal, Ravi Ramamoorthi, Jonathan Barron, and Ren Ng.
\newblock Fourier features let networks learn high frequency functions in low
  dimensional domains.
\newblock In {\em Adv. Neural Inform. Process. Syst.}, 2020.

\bibitem{teed2020raft}
Zachary Teed and Jia Deng.
\newblock Raft: Recurrent all-pairs field transforms for optical flow.
\newblock In {\em Eur. Conf. Comput. Vis.}, 2020.

\bibitem{texler2020interactive}
Ond{\v{r}}ej Texler, David Futschik, Michal Ku{\v{c}}era, Ond{\v{r}}ej
  Jamri{\v{s}}ka, {\v{S}}{\'a}rka Sochorov{\'a}, Menclei Chai, Sergey Tulyakov,
  and Daniel S{\`y}kora.
\newblock Interactive video stylization using few-shot patch-based training.
\newblock {\em ACM Trans. Graph.}, 39(4):73--1, 2020.

\bibitem{tumanyan2023plug}
Narek Tumanyan, Michal Geyer, Shai Bagon, and Tali Dekel.
\newblock Plug-and-play diffusion features for text-driven image-to-image
  translation.
\newblock In {\em IEEE Conf. Comput. Vis. Pattern Recog.}, 2023.

\bibitem{villegas2022phenaki}
Ruben Villegas, Mohammad Babaeizadeh, Pieter-Jan Kindermans, Hernan Moraldo,
  Han Zhang, Mohammad~Taghi Saffar, Santiago Castro, Julius Kunze, and Dumitru
  Erhan.
\newblock Phenaki: Variable length video generation from open domain textual
  description.
\newblock {\em arXiv preprint arXiv:2210.02399}, 2022.

\bibitem{wang2021neus}
Peng Wang, Lingjie Liu, Yuan Liu, Christian Theobalt, Taku Komura, and Wenping
  Wang.
\newblock Neus: Learning neural implicit surfaces by volume rendering for
  multi-view reconstruction.
\newblock {\em arXiv preprint arXiv:2106.10689}, 2021.

\bibitem{wang2021ibrnet}
Qianqian Wang, Zhicheng Wang, Kyle Genova, Pratul~P Srinivasan, Howard Zhou,
  Jonathan~T Barron, Ricardo Martin-Brualla, Noah Snavely, and Thomas
  Funkhouser.
\newblock Ibrnet: Learning multi-view image-based rendering.
\newblock In {\em IEEE Conf. Comput. Vis. Pattern Recog.}, 2021.

\bibitem{wang2022pretraining}
Tengfei Wang, Ting Zhang, Bo Zhang, Hao Ouyang, Dong Chen, Qifeng Chen, and
  Fang Wen.
\newblock Pretraining is all you need for image-to-image translation.
\newblock {\em arXiv preprint arXiv:2205.12952}, 2022.

\bibitem{wang2023zero}
Wen Wang, Kangyang Xie, Zide Liu, Hao Chen, Yue Cao, Xinlong Wang, and Chunhua
  Shen.
\newblock Zero-shot video editing using off-the-shelf image diffusion models.
\newblock {\em arXiv preprint arXiv:2303.17599}, 2023.

\bibitem{wang2019learning}
Xiaolong Wang, Allan Jabri, and Alexei~A Efros.
\newblock Learning correspondence from the cycle-consistency of time.
\newblock In {\em IEEE Conf. Comput. Vis. Pattern Recog.}, 2019.

\bibitem{wang2021real}
Xintao Wang, Liangbin Xie, Chao Dong, and Ying Shan.
\newblock Real-esrgan: Training real-world blind super-resolution with pure
  synthetic data.
\newblock In {\em Int. Conf. Comput. Vis.}, 2021.

\bibitem{wang2022neus2}
Yiming Wang, Qin Han, Marc Habermann, Kostas Daniilidis, Christian Theobalt,
  and Lingjie Liu.
\newblock Neus2: Fast learning of neural implicit surfaces for multi-view
  reconstruction.
\newblock {\em arXiv preprint arXiv:2212.05231}, 2022.

\bibitem{wu2022nuwa}
Chenfei Wu, Jian Liang, Lei Ji, Fan Yang, Yuejian Fang, Daxin Jiang, and Nan
  Duan.
\newblock N{\"u}wa: Visual synthesis pre-training for neural visual world
  creation.
\newblock In {\em Eur. Conf. Comput. Vis.}, 2022.

\bibitem{wu2022tune}
Jay~Zhangjie Wu, Yixiao Ge, Xintao Wang, Weixian Lei, Yuchao Gu, Wynne Hsu,
  Ying Shan, Xiaohu Qie, and Mike~Zheng Shou.
\newblock Tune-a-video: One-shot tuning of image diffusion models for
  text-to-video generation.
\newblock {\em arXiv preprint arXiv:2212.11565}, 2022.

\bibitem{xu2018attngan}
Tao Xu, Pengchuan Zhang, Qiuyuan Huang, Han Zhang, Zhe Gan, Xiaolei Huang, and
  Xiaodong He.
\newblock Attngan: Fine-grained text to image generation with attentional
  generative adversarial networks.
\newblock In {\em IEEE Conf. Comput. Vis. Pattern Recog.}, 2018.

\bibitem{ye2022deformable}
Vickie Ye, Zhengqi Li, Richard Tucker, Angjoo Kanazawa, and Noah Snavely.
\newblock Deformable sprites for unsupervised video decomposition.
\newblock In {\em IEEE Conf. Comput. Vis. Pattern Recog.}, 2022.

\bibitem{yu2021pixelnerf}
Alex Yu, Vickie Ye, Matthew Tancik, and Angjoo Kanazawa.
\newblock pixelnerf: Neural radiance fields from one or few images.
\newblock In {\em IEEE Conf. Comput. Vis. Pattern Recog.}, 2021.

\bibitem{zhang2017stackgan}
Han Zhang, Tao Xu, Hongsheng Li, Shaoting Zhang, Xiaogang Wang, Xiaolei Huang,
  and Dimitris~N Metaxas.
\newblock Stackgan: Text to photo-realistic image synthesis with stacked
  generative adversarial networks.
\newblock In {\em Int. Conf. Comput. Vis.}, 2017.

\bibitem{zhang2023adding}
Lvmin Zhang and Maneesh Agrawala.
\newblock Adding conditional control to text-to-image diffusion models.
\newblock {\em arXiv preprint arXiv:2302.05543}, 2023.

\end{thebibliography}
